\documentclass[10pt,twocolumn,letterpaper]{article}

\usepackage{cvpr}              

\usepackage[margin=1in]{geometry}
\usepackage{booktabs,multirow,makecell,graphicx}
\usepackage{colortbl}
\usepackage{amssymb}
\usepackage{pifont}
\usepackage{float}

\definecolor{colorbest}{RGB}{189,224,200}
\definecolor{colorsecond}{RGB}{226,235,184}
\definecolor{colorthird}{RGB}{254,247,194}
\definecolor{colorNCRNET}{RGB}{198,51,97}
\definecolor{colorEdema}{RGB}{235,185,63}
\definecolor{colorET}{RGB}{101,209,248}

\definecolor{ablationcolorbest}{RGB}{158,202,225}
\definecolor{ablationcolorsecond}{RGB}{198,219,239}
\definecolor{ablationcolorthird}{RGB}{231,240,250}

\newcommand{\abfirst}[1]{\cellcolor{ablationcolorbest}\textbf{#1}}
\newcommand{\absecond}[1]{\cellcolor{ablationcolorsecond}#1}
\newcommand{\abthird}[1]{\cellcolor{ablationcolorthird}#1}

\newcommand{\first}[1]{\cellcolor{colorbest}\textbf{#1}}
\newcommand{\second}[1]{\cellcolor{colorsecond}#1}
\newcommand{\third}[1]{\cellcolor{colorthird}#1}
\newcommand{\filledcirc}{\scalebox{1}{$\bullet$}}
\newcommand{\opencirc}{\scalebox{1}{$\circ$}}
\DeclareRobustCommand{\legendsquare}[1]{%
  \textcolor{#1}{\rule{2ex}{2ex}}%
}

\definecolor{cvprblue}{rgb}{0.21,0.49,0.74}
\usepackage[pagebackref,breaklinks,colorlinks,allcolors=cvprblue]{hyperref}

\def\paperID{12900} 
\def\confName{CVPR}
\def\confYear{2026}

\title{Uni-Encoder Meets Multi-Encoders: Representation Before Fusion for Brain Tumor Segmentation with Missing Modalities}

\author{Peibo Song$^{\text{1}}$, \ Xiaotian Xue$^\text{2}$, \ Jinshuo Zhang$^{\text{1, 3, 4}}$, \ Zihao Wang$^{\text{1}}$, \ Jinhua Liu$^{\text{3, 4}}$, \ Shujun Fu$^\text{1}$, \\ Fangxun Bao$^{\text{1} \ *}$, \ Si Yong Yeo$^{\text{3, 4, 5} \ *}$ \\
$^{\text{1}}$School of Mathematics, Shandong University, \
$^{\text{2}}$GSFS, The University of Tokyo,\\
$^{\text{3}}$MedVisAI Lab, \
$^{\text{4}}$Lee Kong Chian School of Medicine, Nanyang Technological University, \\
$^{\text{5}}$Centre of AI in Medicine, Singapore \\
{\tt\small peibosong@mail.sdu.edu.cn \ fxbao@sdu.edu.cn \ siyong.yeo@ntu.edu.sg}
}

\begin{document}
\maketitle
\renewcommand{\thefootnote}{\fnsymbol{footnote}}
\footnotetext[1]{\textsuperscript{}Corresponding authors.}

\begin{abstract}
    Multimodal MRI offers complementary information for brain tumor segmentation, but clinical scans often lack one or more modalities, which degrades segmentation performance. In this paper, we propose \textbf{UniME} (\textbf{U}ni-Encoder \textbf{M}eets Multi-\textbf{E}ncoders), a two-stage heterogeneous method for brain tumor segmentation with missing modalities that reconciles the trade-offs among fine-grained structure capture, cross-modal complementarity modeling, and exploitation of available modalities. The idea is to decouple representation learning from segmentation via a two-stage heterogeneous architecture. Stage 1 pretrains a single ViT \textbf{Uni-Encoder} with masked image modeling to establish a unified representation robust to missing modalities. Stage 2 adds modality-specific CNN \textbf{Multi-Encoders} to extract high-resolution, multi-scale, fine-grained features. We fuse these features with the global representation to produce precise segmentations. Experiments on BraTS 2023 and BraTS 2024 show that UniME outperforms previous methods under incomplete multi-modal scenarios. The code is available at \href{https://github.com/Hooorace-S/UniME}{https://github.com/Hooorace-S/UniME}.
\end{abstract}
\section{Introduction}\label{sec:intro}
\begin{figure}[t]
    \centering
    \includegraphics[width=1.0\linewidth]{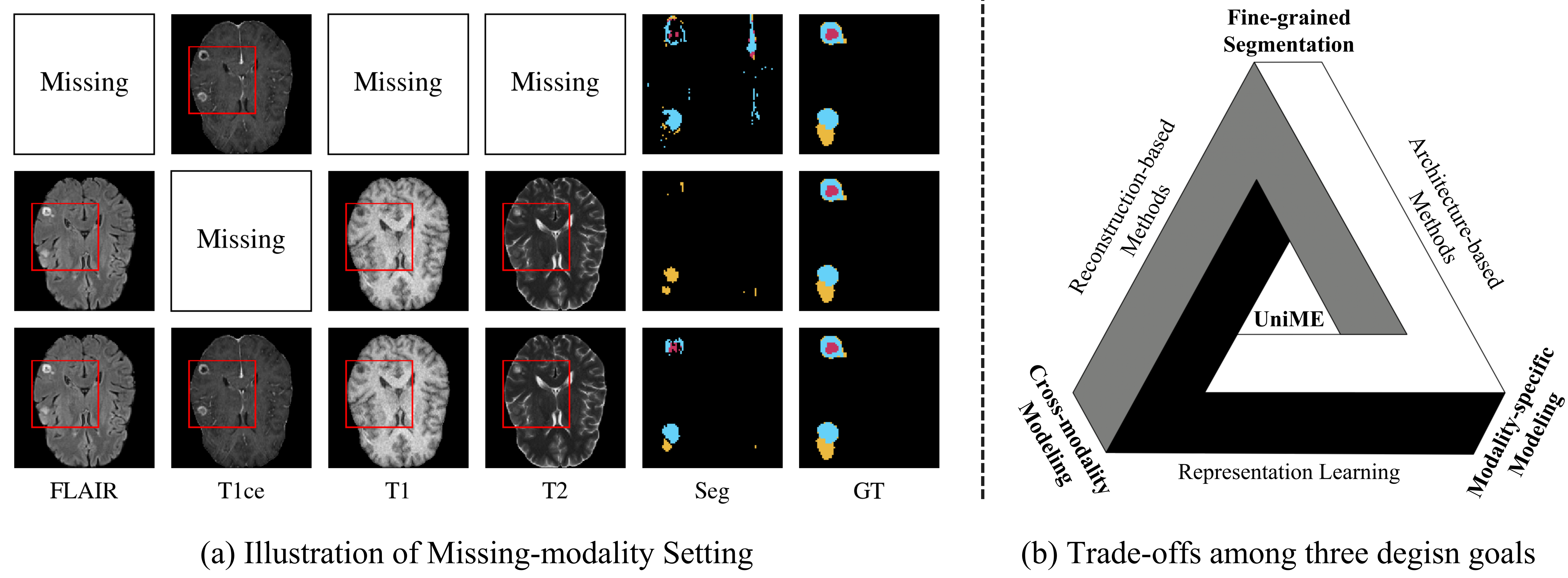}
    \caption{\textbf{(a)~Segmentation under missing MRI modalities.} \legendsquare{colorNCRNET}, \legendsquare{colorET}, and \legendsquare{colorEdema} indicate the necrotic tissue, enhancing tumor, and edema regions, respectively. “Missing” denotes unavailable modality. \textbf{(b)~Trade-offs among three design goals.} Existing methods typically trade off among fine-grained structure capture, cross-modal complementarity modeling, and effective exploitation of available modalities. UniME reconciles these trade-offs by introducing a two-stage heterogeneous design, which decouples representation learning from segmentation.}
    \label{fig:background}
\end{figure}

Accurate brain tumor segmentation from multi-modal MRI is clinically important~\cite{bakas2018identifying}. Since different modalities provide complementary information, high-quality segmentation requires capturing fine anatomical details and effectively leveraging cross-modal complementarity~\cite{tseng2017joint, zhang2020exploring}. However, in clinical practice, one or more MRI modalities are often missing~\cite{krupa2015artifacts}. For example, T2-weighted MRI is usually unavailable due to its susceptibility to artifacts~\cite{graves2013body}. As illustrated in Fig.~\ref{fig:background}(a), such incomplete modality settings degrade segmentation performance~\cite{chen2023confidence}, raising two challenges: \textit{(i) how to model cross-modal complementarity to mitigate the impact of missing modalities, and (ii) how to fully exploit the information present within the available modalities.}

Recent methods mainly fall into two families. The first aims to maximize cross-modal complementarity, via modality synthesis~\cite{lee2020assessing, zhou2020hi, yu2022mousegan++, yang2023learning, kim2024adaptive} or knowledge distillation~\cite{wang2021acn, azad2022smu, qiu2023scratch, wei2023mmanet}. However, synthesis usually fails to recover fine details, and distillation pipelines are often complex and difficult to scale. The second family redesigns architectures to better use the available modalities~\cite{havaei2016hemis, ding2021rfnet, zhang2022mmformer, shi2023mftrans,  wang2023multi, zhang2025incomplete, pipoli2025fuse}, but the limited receptive fields and high computational cost hinder the capture of global, cross-modal semantics. As shown in Fig.~\ref{fig:background}(b), an ideal method should capture fine-grained structures, model cross-modal complementarity, and effectively exploit the available modalities, but achieving these goals simultaneously remains challenging.

The rise of self-supervised representation learning offers a new perspective~\cite{caron2021emerging, he2022masked, xie2022simmim, wei2022masked, fang2024eva}. Large-scale self-supervised Vision Transformer (ViT) models have been shown to effectively capture global semantics and model cross-modal relationships~\cite{oquab2023dinov2, assran2023self, kirillov2023segment, ravi2024sam, fang2024eva, kerssies2025your, simeoni2025dinov3}. This suggests a new route to capture global semantics and model  cross-modal complementarity. However, in medical image segmentation, a pure ViT network without convolutional inductive biases often fails to meet the pixel-level precision required to capture fine anatomical structures~\cite{wald2025primus}. Moreover, relying solely on a unified encoder is insufficient to fully exploit modality-specific information, limiting segmentation accuracy under incomplete modality scenarios.

To this end, we propose \textbf{UniME}, a two-stage heterogeneous segmentation method that explicitly decouples representation learning from segmentation, effectively modeling cross-modal complementarity through masked self-supervised learning, and fully utilizing available modalities while capturing fine-grained structures via a heterogeneous architecture design. In \textbf{Stage 1}, we pretrain a single ViT \textbf{Uni-Encoder} with masked self-supervision that combines modality-level and patch-level masking to learn a unified representation robust to missing modalities. To improve semantic generalization under limited data, the Uni-Encoder incorporates spatial priors through effective positional embedding strategies. In \textbf{Stage 2}, we add parallel modality-specific CNN \textbf{Multi-Encoders} to extract high-resolution, multi-scale, fine-grained features, which enrich Uni-Encoder features with multi-scale detail. To ensure smooth fine-tuning, we apply layer-wise learning rate decay (LLRD), which helps preserve the pretrained semantic knowledge captured by the Uni-Encoder. In summary, our main contributions are:

\begin{itemize}
    \item We introduce \textbf{UniME}, to our knowledge, the \textit{first} two-stage heterogeneous method for brain tumor segmentation with missing modalities, effectively reconciling the trade-offs among fine-grained structure capture, cross-modal complementarity modeling, and effective exploitation of available modalities.
    \item To model cross-modal complementarity, UniME pretrains a Uni-Encoder with masked self-supervision to obtain unified representations robust to absent modalities.
    \item To exploit available modalities and capture fine-grained details, UniME employs a heterogeneous architecture that introduces Multi-Encoders to enrich the Uni-Encoder with multi-scale, fine-grained features.
    \item Experiments on BraTS 2023 and BraTS 2024 validate UniME’s effectiveness, particularly in scenarios with missing modalities.
\end{itemize}
\section{Related Work}
\label{sec:related_work}
\noindent\textbf{Incomplete Multimodal Tumor Segmentation.}~~Incomplete data are common in practice~\cite{zhao2019data, wang2020lt, tran2017missing}. In this work, we study brain tumor segmentation with missing modalities. Compared with standard brain tumor segmentation~\cite{yeo2011level,ronneberger2015u, yang2016cardiac, cao2022swin, chen2024transunet, hatamizadeh2021swin, wald2025primus}, the missing-modality setting is both more practical and more challenging than the standard setting. Limitations in modality synthesis~\cite{lee2020assessing, zhou2020hi, yu2022mousegan++, yang2023learning, kim2024adaptive} and in knowledge distillation~\cite{wang2021acn, azad2022smu, qiu2023scratch, wei2023mmanet} have shifted attention to architectural designs that better use the available modalities. Havaei et al.~\cite{havaei2016hemis} introduced HeMIS, which uses four parallel encoders and a shared decoder for multimodal fusion. This established a baseline architecture adopted by much subsequent work. Later studies largely retained the HeMIS framework and augmented it with stronger modules or fusion strategies~\cite{ding2021rfnet, zhang2022mmformer, shi2023mftrans, wang2023multi, zhang2025incomplete, pipoli2025fuse}. Despite these advances, most methods follow the HeMIS paradigm: extract features with multiple independent CNN encoders and fuse them downstream. Even with Transformer-based or other global context fusion~\cite{zhang2022mmformer, shi2023mftrans, ding2021rfnet, zhang2025incomplete, pipoli2025fuse}, the upstream CNN encoders impose local receptive field limits. This dependency forms a performance ceiling for cross-modal feature fusion and is a key bottleneck to further progress.

\noindent\textbf{Representation Learning.}~~In recent years, self-supervised representation learning has reshaped computer vision~\cite{caron2021emerging, he2022masked, xie2022simmim, wei2022masked, fang2024eva}. Masked pretraining methods show that a single ViT encoder can learn transferable representations from unlabeled data~\cite{he2022masked, xie2022simmim, wei2022masked, fang2024eva}. Moreover, recent studies show that large-scale, extensively pretrained ViTs can both capture high-level semantics and model cross-modal relationships~\cite{oquab2023dinov2, assran2023self, kirillov2023segment, ravi2024sam, fang2024eva, kerssies2025your, simeoni2025dinov3}. For brain tumor segmentation with missing modalities, several methods also adopt two-stage masked modeling. Liu et al.~\cite{liu2023m3ae} proposed M$^{3}$AE, which uses a symmetric U-Net for both pretraining and fine-tuning. However, its full CNN decoder and skip connections diminish the encoder’s effectiveness in representation learning~\cite{xie2022simmim, zhang2024residual}. Similarly, Zhang et al.~\cite{zhang2025m2seg} introduced M$^{2}$SegMamba, which applies semantic-level masking while retaining a multi-encoder architecture. Nevertheless, its reliance on multiple CNN encoders and a full decoder architecture can hinder effective feature alignment across different modalities. These choices reflect 3D medical segmentation’s emphasis on spatial priors and pixel-wise accuracy under limited data. As a result, the advantages of ViT may diminish in data-limited settings. Recent evidence also shows that Transformer-based or hybrid models do not consistently outperform U-Net and its variants in 3D medical image segmentation~\cite{wald2025primus}.

By contrast, UniME explicitly decouples representation learning from segmentation through a two-stage heterogeneous design, better leveraging ViT architectures and self-supervised pretraining for MRI data with missing modalities.
\section{Method}
\begin{figure*}[htbp]
    \centering
    \includegraphics[width=0.85\linewidth]{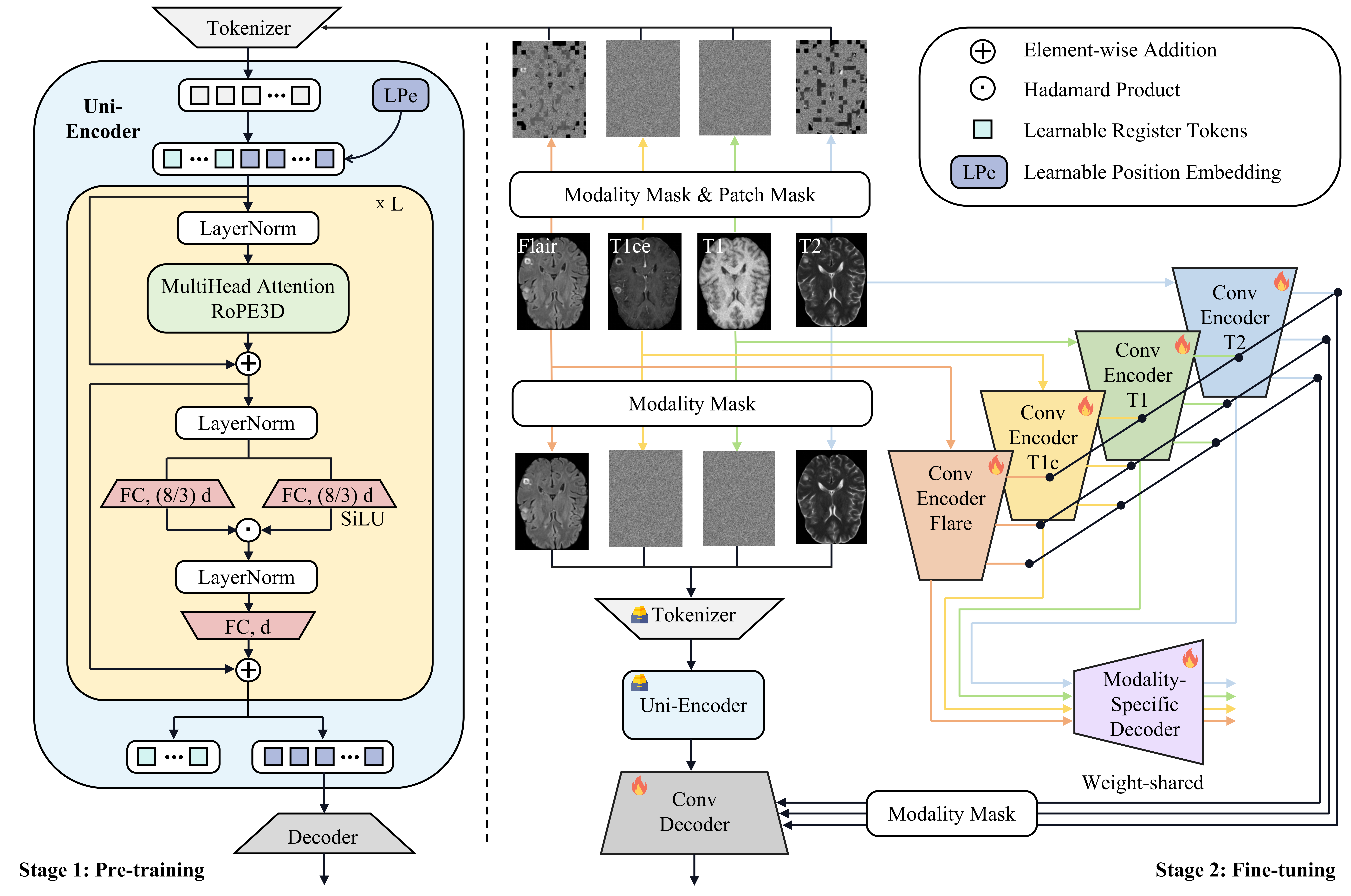}
    \caption{\textbf{UniME overview.} Stage 1 pretrains a single ViT Uni-Encoder with masked self-supervision and a lightweight auxiliary decoder that is discarded after pretraining. Stage 2 introduces parallel modality-specific encoders and fuses multi-scale features for segmentation. \includegraphics[height=1em]{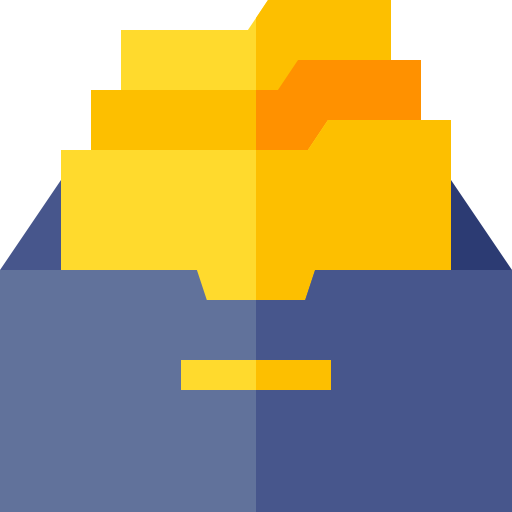} marks modules initialized from pretraining weights, while \includegraphics[height=1em]{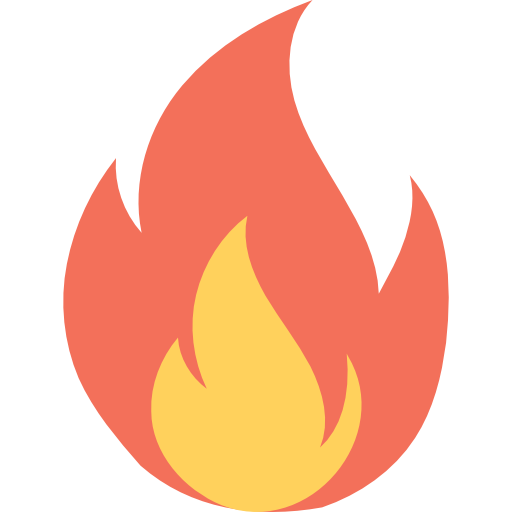} indicates modules trained from scratch.}
    \label{fig:framework}
\end{figure*}

For brain tumor segmentation with missing modalities, an ideal method should capture fine-grained structures, model cross-modal complementarity, and effectively exploit the available modalities. However, as illustrated in Fig.~\ref{fig:background}(b), existing approaches rarely achieve all three goals simultaneously. To effectively address these challenges, we propose a two-stage heterogeneous design principle, which decouples visual representation learning from brain tumor segmentation, thus providing an effective solution to this problem. Based on this principle, we introduce UniME. Fig.~\ref{fig:framework} provides an overview of UniME. Next, we shall outline the two-stage heterogeneous design and then detail each stage.

\subsection{Two-Stage Heterogeneous Design}

In Stage 1, we pretrain a single ViT Uni-Encoder with masked self-supervised learning to capture global semantics and to model cross-modal complementarity. To mitigate data scarcity, the Uni-Encoder combines learnable positional embeddings with 3D rotary position embeddings, which embed spatial priors explicitly. Unlike previous work~\cite{liu2023m3ae, zhang2025m2seg}, we attach a lightweight auxiliary decoder during pretraining and discard it afterward. This design keeps representation learning encoder-centric and avoids reliance on heavy decoders for reconstruction.

In Stage 2, a heterogeneous network fuses semantic features from the Uni-Encoder with fine-grained structural features extracted by parallel modality-specific encoders. Cross-level skip connections enrich semantic representations with fine-grained details. A weight-shared decoder produces a prediction for each modality-specific encoder and enables separate gradient flows per modality. This stabilizes multi-scale feature learning. We also use the LLRD strategy to preserve pretrained high-quality semantic representations and to improve stability and generalization during fine-tuning.

\subsection{Stage 1: Uni-Encoder Pretraining}
In this subsection, we describe pretraining of the Uni-Encoder with masked self-supervised learning. We begin with the masking strategy and then detail the patch tokenizer, the Uni-Encoder architecture, and the loss function. The left part of Fig.~\ref{fig:framework} provides an overview of Stage 1.

\noindent \textbf{Masking Strategy.}~~We adopt the masking strategy of M$^{3}$AE~\cite{liu2023m3ae}, which combines modality-level random masking with patch-level random masking. Let $\mathbf{X} \in \mathbb{R}^{K \times D \times H \times W}$ be the input with $K$ modalities, and let $\mathcal{M} = \{\text{FLAIR}, \text{T1ce}, \text{T1}, \text{T2}\}$ denote the modality set. Let $P$ be the patch size, and $N=(DHW) / P^{3}$ the number of patch tokens for the channel-stacked volume. We define modality-level and patch-level binary mask indicators $\delta_m$ and $\eta_{m, i}$ as:
\begin{equation}\label{eq:mask_indicator}
    \begin{aligned}
        \delta_m    & \sim \text{Bernoulli}(1 - p_m),
        \\
        \eta_{m, i} & \sim \text{Bernoulli}(1 - q_m),
    \end{aligned}
\end{equation}
where $m \in \mathcal{M}$ and $i \in \{1, \dots, N\}$. $\text{Bernoulli} \left( \cdot \right)$ denotes the Bernoulli distribution. The parameter $p_{m}$ is the probability of masking modality $m$, and $q_{m}$ is the probability of masking patch $i$ of modality $m$. To ensure at least one modality is available, we enforce $\sum_{m}\delta_m \ge 1$ and resample until it holds. Consequently, the masking strategy is formulated as:
\begin{equation}\label{eq:masking_strategy}
    \mathbf{X}_{m, i} \leftarrow \gamma_{m, i}\mathbf{X}_{m,i} + (1 - \gamma_{m,i})\texttt{Mask}_{m,i},
\end{equation}
where $\gamma_{m,i}=\delta_m \cdot \eta_{m,i}$ is the joint masking indicator. The term $\mathbf{X}_{m,i}$ is the $i$-th patch of modality $m$, and $\texttt{Mask}_{m,i}$ denotes the learnable mask token at that position. In practice, we set $p_{m} = 0.5$, which balances coverage across non-empty modality subsets. The parameter $q_m$ is treated as a hyperparameter. Its effect is examined in the ablation studies. The resulting masked volume mixes visible patches and mask tokens and is then tokenized by the patch tokenizer.

\noindent \textbf{Patch Tokenizer.}~~The tokenizer partitions the masked input $\mathbf{X}$ from Eq.~\ref{eq:masking_strategy} into token sequences. To maintain consistency between masking and tokenization, we use the same patch size $P$ as in the masking strategy. Let $d_{\text{embed}}$ denote the token embedding dimension. We implement the tokenizer as a 3D convolution with kernel size and stride of $P$, and with $d_{\text{embed}}$ as the output channels. To capture fine-grained structures in medical images~\cite{salpea2022medical, xiao2023transformers}, we use $P=8$ rather than the typical $P=16$ for natural images~\cite{dosovitskiy2020image}. Although $P=8$ increases sequence length, we concatenate modalities along the channel axis rather than the token axis, which keeps the final sequence length comparable to previous work~\cite{zhang2022mmformer, pipoli2025fuse, zhang2025incomplete, zhang2025m2seg}. Additionally, recent efficient attention algorithms~\cite{dao2022flashattention, dao2023flashattention2} further offset the cost of longer sequences.

\noindent \textbf{Uni-Encoder Architecture.}~~The Uni-Encoder builds upon the foundational ViT architecture~\cite{vaswani2017attention, dosovitskiy2020image} and integrates recent advancements~\cite{dauphin2017language,shazeer2020glu,su2024roformer,gervet2023act3d,fang2024eva,wald2025primus}. First, following Darcet et al.~\cite{darcet2023vision}, we introduce register tokens to improve robustness in masked self-supervised pretraining. Second, because spatial priors are critical for global sequence models~\cite{hatamizadeh2021swin,baron20232,zhu2024vision}, especially with limited medical data, we combine learnable positional embeddings (LPe) and 3D rotary position embeddings (RoPE)~\cite{su2024roformer,gervet2023act3d,wald2025primus} to encode position consistently. LPe offers flexible, learnable offsets, whereas 3D RoPE is applied in each multi-head attention layer to embed relative spatial relationships consistently. Third, we adopt SwiGLU in the feed-forward network (FFN) modules, given its effectiveness in large language models~\cite{shazeer2020glu,liu2024deepseek,dubey2024llama,yang2025qwen3}. Following EVA-02~\cite{fang2024eva}, we apply layer normalization both before and after the FFN to improve stability and convergence.

Let $\mathbf{S}$ denote the tokenizer output. We first add LPe and append the register tokens:
\begin{equation}\label{eq:}
    \mathbf{S}^{\left( 0 \right)} = \left[\mathbf{S} + \mathbf{E}_{\text{LPe}}, \mathbf{E}_{\text{Reg}}\right] \in \mathbb{R}^{\left( N + N_{\text{reg}} \right) \times d_{\text{embed}}},
\end{equation}
where $\mathbf{E}_{\text{LPe}}\in\mathbb{R}^{N\times d_{\text{embed}}}$ denotes the LPe tokens, and $\mathbf{E}_{\text{Reg}}\in\mathbb{R}^{N_{\text{reg}}\times d_{\text{embed}}}$ denotes the registers with $N_{\text{reg}}$ tokens. Then, the Uni-Encoder applies $L$ Transformer layers as follows:
\begin{equation}\label{eq:uniencoder_layer}
    \begin{aligned}
        \widehat{\mathbf{S}}^{\left( l \right)}  & = \mathbf{S}^{\left( l - 1 \right)} + \text{MHSA}_{\text{3D RoPE}} ( \text{LN}( \mathbf{S}^{( l - 1 )} )),
        \\
        \overline{\mathbf{S}}^{\left( l \right)} & = \text{LN} ( \text{SwiGLU} ( \text{LN} ( \widehat{\mathbf{S}}^{\left( l \right)} ) ) ),
        \\
        \mathbf{S}^{\left( l \right)}            & = \widehat{\mathbf{S}}^{\left( l \right)} + \overline{\mathbf{S}}^{\left( l \right)},
    \end{aligned}
\end{equation}
where $\mathrm{MHSA}_{\text{3D RoPE}} $ denotes multi-head attention with 3D RoPE. After $L$ layers, we discard the register tokens and obtain the final representation. We use $L=16$, $d_{\text{embed}}=864$, $N_{\text{reg}}=4$, and 12 attention heads by default. In ablation studies, we also explore how varying model size via different combinations of $L$, $d_{\text{embed}}$, $N_{\text{reg}}$, and attention heads affects performance.

\noindent \textbf{Loss Function.}$\quad$To encourage semantic learning of the Uni-Encoder, we attach a lightweight auxiliary decoder during pretraining. The decoder consists of three 3D transposed convolution upsampling blocks, each followed by layer normalization and GELU. Let $\widehat{\mathbf{X}}$ denote the reconstruction. The loss is defined by:
\begin{equation}\label{eq:stage1_loss}
    \mathcal{L}_{\text{rec}} = \| \mathbf{X} - \widehat{\mathbf{X}} \|_{2}^{2} + \gamma \| \texttt{Mask} \|_{2},
\end{equation}
where $\gamma = 0.005$ by default, \texttt{Mask} denotes the learnable mask tokens and $\| \cdot \|_2$ is the $L^2$ norm. Since modality drop can hide an entire modality channel, and high patch-mask rates often leave no location with a complete set of modalities, we therefore compute reconstruction over the entire input and output volume to encourage richer cross-modal representations, which differs from the masked-region loss used in MAE~\cite{he2022masked}.

\subsection{Stage 2: Network Fine-tuning}
We now describe Stage 2, including the segmentation network architecture, the LLRD strategy, and the loss function design. The right part of Fig.~\ref{fig:framework} provides an overview of Stage 2.

\noindent \textbf{Segmentation Network.}~~The segmentation network uses two parallel feature-extraction paths. Let $\mathbf{X} \in \mathbb{R}^{K \times D \times H \times W}$ be the input with $K$ modalities. The first path randomly masks modalities in $\mathbf{X}$ and processes the result with the Stage 1 pretrained Uni-Encoder, which yields the multimodal semantic representation $\mathbf{F}_{\mathrm{Uni}}$. The second path extracts features for each modality with a specific CNN encoder. We first split $\mathbf{X}$ along the modality axis into $\{ \mathbf{X}_{m}: m \in \mathcal{M}\}$. Each $\mathbf{X}_{m}$ is then processed by its encoder $\mathbf{Enc}_m$. The encoder $\mathbf{Enc}_m$ follows the U-Net encoder with four stages, each comprising three convolutional blocks. Each block contains a 3D convolutional layer, instance normalization, and a GELU activation. To downsample progressively, the first block of each stage uses stride~2, except at stage zero where the stride is~1. Within each stage, a residual connection adds the output of the first block to the post-activation output of the third block. The stage channel widths are 16, 32, 64, and 128.

For the first three stages, modality-specific features corresponding to unavailable modalities are masked before fusion. We then concatenate all the features along channels and feed them to a multimodal fusion block. Each fusion block consists of two 3D convolutional blocks and an Efficient Channel Attention (ECA) module~\cite{wang2020eca}. The first block projects and fuses channels. The second block extracts higher-level fused features. The ECA module adaptively emphasizes discriminative channels. Let $\mathbf{F}^{(0)}$, $\mathbf{F}^{(1)}$, and $\mathbf{F}^{(2)}$ denote the multi-scale features produced by these fusion blocks. We \textit{exclude} the deepest modality-specific features at the \textit{last} stage from fusion because the Uni-Encoder already provides a high-quality multimodal semantic representation. We also process $\mathbf{F}_{\mathrm{Uni}}$ with a fusion block to integrate all modality information. This block reduces the channel dimension and yields $\mathbf{F}_{\mathrm{main}}$. Finally, a symmetric U-Net-style CNN decoder takes $\mathbf{F}_{\mathrm{main}}$ and uses $\mathbf{F}^{(0)}$, $\mathbf{F}^{(1)}$, and $\mathbf{F}^{(2)}$ as skip connections, producing the primary output $\mathbf{O}_{\mathrm{main}}$, along with intermediate deep supervision outputs.

Additionally, to stabilize the gradient flows of each modality-specific encoder under random modality masking, we introduce a decoder with shared weights. This decoder shares weights across modalities, takes multi-scale features \textit{including} the deepest level from each modality-specific encoder, and produces outputs for auxiliary supervision.

\noindent \textbf{LLRD Strategy.}~~We employ the LLRD strategy to stabilize training and preserve pretrained semantic representations of the Uni-Encoder. More specifically, let \texttt{lr} be the base learning rate. Uni-Encoder parameters in layer $l$ are scaled by a layer-wise decay factor $\omega^{L-l}$:
\begin{equation}\label{eq:llrd}
    \texttt{lr}_{l} = \texttt{lr} \cdot \omega^{L - l}, \quad l = 1, 2, \dots, L.
\end{equation}
Thus, lower layers receive smaller learning rates that preserve general features, and deeper layers receive larger rates that adapt to task-specific requirements. The ablation studies also investigate the impact of varying the decay factor $\omega$, comparing frozen versus directly fine-tuned Uni-Encoder parameters.

\noindent \textbf{Loss Function.}$\quad$Following the previous methods~\cite{ding2021rfnet,zhang2022mmformer,liu2023m3ae, zhang2025m2seg,pipoli2025fuse, zhang2025incomplete}, we use the following loss function:
\begin{equation}\label{eq:stage2_loss}
    \mathcal{L}_{\text{total}} = \mathcal{L}_{\text{main}} + \mathcal{L}_{\text{aux}} + \mathcal{L}_{\text{deep}}.
\end{equation}
Each component combines the Dice loss with the weighted cross-entropy loss in order to address class imbalance in multi-class segmentation. $\mathcal{L}_{\text{main}}$ is computed from the main output $\mathbf{O}_{\text{main}}$. $\mathcal{L}_{\mathrm{aux}}$ sums the losses over the modality-specific auxiliary outputs. $\mathcal{L}_{\mathrm{deep}}$ sums the losses from the deep supervision outputs.

\section{Experiments}

\subsection{Dataset and implementation details}
Experiments use the BraTS 2023~\cite{adewole2023brain} and BraTS 2024~\cite{de20242024} datasets from the Multimodal Brain Tumor Segmentation Challenge, containing 1251 and 1350 cases, respectively, with publicly available ground truth labels. BraTS 2023 and BraTS 2024 represent different clinical stages, namely pre-operative and post-operative settings, respectively. We use a split of 70\% for training, 10\% for validation, and 20\% for testing. The model achieving the best metric on the validation set is selected for evaluation on the test set. Each case includes four MRI modalities: T1, T1ce, T2, and FLAIR. For BraTS 2023, these modalities capture complementary characteristics of three subregions: enhancing tumor, peritumoral edema, and the necrotic or non-enhancing tumor core. These components form three nested subregions: the whole tumor (WT), the tumor core (TC), and the enhancing tumor (ET). We quantify segmentation performance with the Dice Similarity Coefficient (DSC) and 95th percentile Hausdorff Distance (HD95).

Our model is implemented in PyTorch 2.8.0 and trained on an NVIDIA RTX 6000 Blackwell GPU. Training involves randomly cropping volumes to $96 \times 96 \times 96$ voxels and applying data augmentation through random rotations, flips, and intensity adjustments. Both pretraining and fine-tuning use a batch size of 4, running for 600 epochs at 250 iterations each. The optimizer uses AdamW with weight decay $1 \times 10^{-4}$. The learning rate follows a stepwise schedule: it starts at $1 \times 10^{-5}$, increases linearly to $3 \times 10^{-4}$ over the first 5\% of steps, and then decays with a cosine profile back to $1 \times 10^{-6}$ by the end of training. GPU memory usage during training is approximately 23.80 GiB. Further details, including label definition of BraTS 2024, data preprocessing, inference methods, and model complexity, are provided in the supplementary materials.

\subsection{Comparisons with the State-of-the-art}
We compare UniME with 9 state-of-the-art methods on different cases with missing modalities, including HeMIS~\cite{havaei2016hemis}, U-HVED~\cite{dorent2019hetero}, RFNet~\cite{ding2021rfnet}, mmFormer~\cite{zhang2022mmformer}, M$^{3}$AE~\cite{liu2023m3ae}, M$^{2}$FTrans~\cite{shi2023mftrans}, LS3M~\cite{zhang2025incomplete}, IM-Fuse~\cite{pipoli2025fuse}, and M$^{2}$SegMamba~\cite{zhang2025m2seg}. For fair comparison, all methods are trained under their recommended hyperparameters within the same dataset split.

\begin{figure*}[htbp]
    \centering
    \includegraphics[width=1\linewidth]{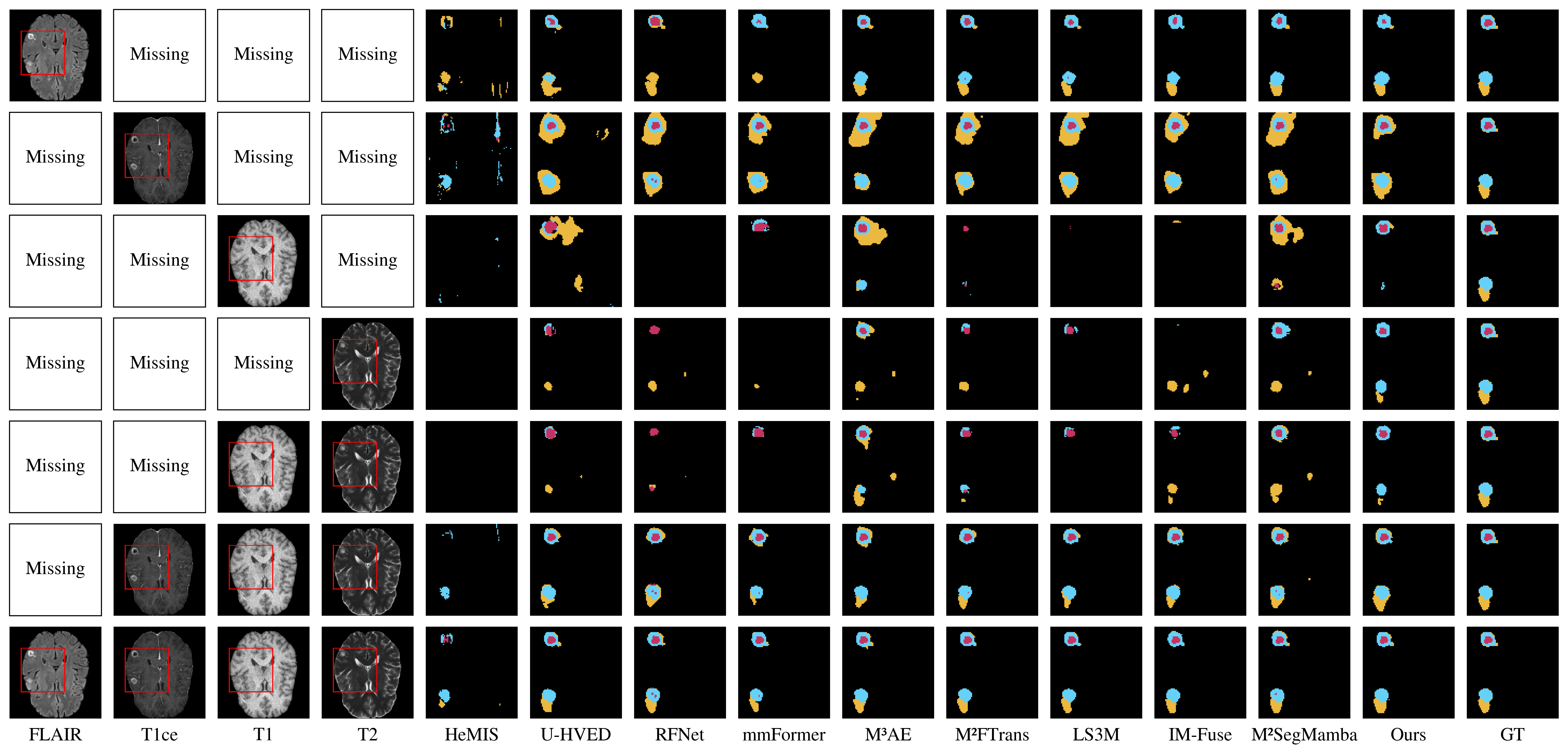}
    \caption{\textbf{Visualization of segmentation results on BraTS 2023.} The segmentation masks predicted by different methods for various modality combinations (FLAIR, T1ce, T1, T2) are presented within the red-framed regions of the input images on the left. Each row represents a distinct patient case with different modality availabilities. The columns demonstrate segmentation outputs from multiple approaches, compared against the ground truth (GT). \legendsquare{colorNCRNET}, \legendsquare{colorET}, and \legendsquare{colorEdema} indicate the necrotic tissue, enhancing tumor, and edema regions, respectively. “Missing” denotes unavailable modalities.}
    \label{fig:visualization_comparison}
\end{figure*}

As shown in Tab.~\ref{tab:dsc23} and Tab.~\ref{tab:dsc24}, our method achieves the superior results across most missing-modality combinations. On BraTS 2023, our method improves the average DSC by 1.40\%, 1.53\%, and 2.36\% for WT, TC, and ET compared to the second-best method. On BraTS 2024, the improvements are 1.84\% for WT, 1.96\% for TC, and 2.93\% for ET. Additionally, Fig.~\ref{fig:visualization_comparison} shows that our method yields more accurate segmentation results, particularly when fewer modalities are available. Comparisons based on HD95 are provided in the supplementary material.

\begin{table*}[htbp]
    \centering
    \scriptsize
    \renewcommand{\arraystretch}{1}
    \setlength{\tabcolsep}{4.3pt}
    \begin{tabular}{c|c|*{15}{c}|c}
        \toprule
        \multirow{4}{*}{\textbf{Type}} & \multicolumn{1}{c|}{Flair} & \opencirc      & \opencirc      & \opencirc      & \filledcirc    & \opencirc      & \opencirc      & \filledcirc    & \opencirc      & \filledcirc    & \filledcirc    & \filledcirc    & \filledcirc    & \filledcirc    & \opencirc      & \filledcirc    & \multirow{4}{*}{\textbf{Avg.}} \\
                                       & T1                         & \opencirc      & \opencirc      & \filledcirc    & \opencirc      & \opencirc      & \filledcirc    & \filledcirc    & \filledcirc    & \opencirc      & \opencirc      & \filledcirc    & \filledcirc    & \opencirc      & \filledcirc    & \filledcirc    &                                \\
                                       & T1ce                       & \opencirc      & \filledcirc    & \opencirc      & \opencirc      & \filledcirc    & \filledcirc    & \opencirc      & \opencirc      & \opencirc      & \filledcirc    & \filledcirc    & \opencirc      & \filledcirc    & \filledcirc    & \filledcirc    &                                \\
                                       & T2                         & \filledcirc    & \opencirc      & \opencirc      & \opencirc      & \filledcirc    & \opencirc      & \opencirc      & \filledcirc    & \filledcirc    & \opencirc      & \opencirc      & \filledcirc    & \filledcirc    & \filledcirc    & \filledcirc    &                                \\
        \midrule
        \multirow{10}{*}{\textbf{WT}}  & HeMIS                      & 35.09          & 26.36          & 11.70          & 68.97          & 60.26          & 39.89          & 73.06          & 56.60          & 75.02          & 75.08          & 77.58          & 77.68          & 79.04          & 60.82          & 79.08          & 59.75                          \\
                                       & U-HEVD                     & 81.28          & 71.55          & 70.14          & 82.02          & 85.65          & 76.43          & 87.49          & 85.16          & 86.96          & 87.33          & 88.61          & 88.81          & 89.25          & 86.64          & 89.74          & 83.80                          \\
                                       & RFNet                      & 85.61          & 77.98          & 76.41          & 89.07          & 88.08          & 80.87          & 90.92          & 87.41          & 90.81          & 91.42          & 91.79          & 91.37          & 91.82          & 88.50          & 91.98          & 87.60                          \\
                                       & mmFormer                   & 85.89          & 77.07          & 77.07          & 89.31          & 88.03          & 80.32          & 90.65          & 87.46          & 90.78          & 91.27          & 91.51          & 91.18          & 91.65          & 88.37          & 91.72          & 87.48                          \\
                                       & M$^{3}$AE                  & \third{87.60}  & \second{80.83} & \second{80.25} & \second{91.03} & 88.99          & 82.15          & \third{91.73}  & \second{88.96} & \third{91.84}  & \third{92.02}  & \third{92.20}  & \third{92.16}  & \third{92.54}  & \third{89.46}  & \third{92.69}  & \third{88.96}                  \\
                                       & M$^{2}$FTrans              & 86.96          & 79.58          & 78.28          & 89.72          & 88.87          & 82.10          & 90.97          & 88.35          & 91.22          & 91.52          & 91.69          & 91.53          & 92.05          & 89.12          & 92.10          & 88.27                          \\
                                       & LS3M                       & 86.96          & 78.41          & 77.46          & 89.94          & 88.77          & 81.52          & 91.20          & 88.14          & 91.37          & 91.63          & 92.01          & 91.72          & 92.15          & 89.12          & 92.25          & 88.18                          \\
                                       & IM-Fuse                    & 87.08          & \third{80.02}  & 79.48          & \third{90.87}  & \third{89.03}  & \second{82.83} & 91.39          & 88.62          & 91.74          & 91.98          & 92.06          & 91.90          & 92.33          & 89.39          & 92.37          & 88.74                          \\
                                       & M$^{2}$SegMamba            & \second{87.76} & 79.03          & \third{79.81}  & 90.72          & \second{89.79} & \third{82.38}  & \second{91.78} & \third{88.81}  & \second{91.86} & \second{92.39} & \second{92.40} & \second{92.29} & \second{93.03} & \second{89.71} & \second{93.02} & \second{88.98}                 \\
                                       & Ours                       & \first{89.19}  & \first{84.52}  & \first{83.11}  & \first{91.94}  & \first{90.35}  & \first{85.73}  & \first{92.49}  & \first{89.95}  & \first{92.68}  & \first{93.05}  & \first{93.04}  & \first{92.83}  & \first{93.15}  & \first{90.55}  & \first{93.13}  & \first{90.38}                  \\
        \hline
        \multirow{10}{*}{\textbf{TC}}  & HeMIS                      & 9.47           & 22.71          & 1.36           & 29.77          & 54.39          & 46.33          & 35.41          & 16.18          & 36.57          & 62.97          & 66.05          & 34.93          & 67.31          & 56.75          & 66.20          & 40.43                          \\
                                       & U-HEVD                     & 57.06          & 75.68          & 49.04          & 56.91          & 81.76          & 78.93          & 65.55          & 61.21          & 64.64          & 81.51          & 82.67          & 66.60          & 83.31          & 82.65          & 83.74          & 71.42                          \\
                                       & RFNet                      & 66.19          & 85.30          & 64.53          & 68.83          & 87.01          & 86.53          & 72.86          & 70.61          & 73.47          & 87.31          & 88.22          & 74.30          & 87.77          & 87.24          & 87.85          & 79.20                          \\
                                       & mmFormer                   & 65.77          & 84.74          & 65.00          & 64.65          & 87.25          & 87.16          & 70.62          & 70.19          & 69.72          & 87.76          & 88.31          & 72.91          & 87.63          & 87.62          & 88.01          & 78.49                          \\
                                       & M$^{3}$AE                  & \second{74.17} & \second{87.98} & \second{71.57} & \second{74.10} & \second{89.66} & 88.47          & \second{76.88} & \second{75.68} & \second{77.98} & \second{89.82} & \second{90.16} & \second{78.48} & \second{90.17} & 89.30          & \second{90.29} & \second{82.98}                 \\
                                       & M$^{2}$FTrans              & \third{72.21}  & \third{87.82}  & 66.89          & 70.87          & 88.77          & 88.92          & 74.56          & 73.98          & 75.87          & 89.09          & 89.40          & 76.38          & 89.14          & 88.94          & 89.43          & 81.49                          \\
                                       & LS3M                       & 69.05          & 87.14          & 67.16          & 71.08          & 89.02          & \second{89.46} & 74.67          & 72.94          & 75.61          & 89.02          & \third{89.80}  & 76.43          & 89.37          & \second{90.05} & 89.81          & 81.38                          \\
                                       & IM-Fuse                    & 68.09          & 87.28          & 68.45          & 71.31          & 87.89          & 88.48          & 74.02          & 71.63          & 73.64          & 88.53          & 89.08          & 74.95          & 88.36          & 88.46          & 88.72          & 80.59                          \\
                                       & M$^{2}$SegMamba            & 71.48          & 87.12          & \third{70.17}  & \third{72.32}  & \third{89.43}  & \third{89.10}  & \third{75.97}  & \third{74.43}  & \third{75.88}  & \third{89.16}  & 89.75          & \third{77.49}  & \third{89.73}  & \third{89.69}  & \third{89.98}  & \third{82.11}                  \\
                                       & Ours                       & \first{75.71}  & \first{90.06}  & \first{73.42}  & \first{77.01}  & \first{91.19}  & \first{90.30}  & \first{78.87}  & \first{77.19}  & \first{79.20}  & \first{90.72}  & \first{90.79}  & \first{79.89}  & \first{91.09}  & \first{91.09}  & \first{91.11}  & \first{84.51}                  \\
        \hline
        \multirow{10}{*}{\textbf{ET}}  & HeMIS                      & 4.96           & 30.62          & 5.91           & 18.92          & 57.34          & 46.31          & 18.89          & 7.60           & 21.46          & 65.83          & 68.54          & 18.76          & 68.56          & 60.23          & 67.74          & 37.44                          \\
                                       & U-HEVD                     & 35.95          & 71.10          & 22.41          & 36.63          & 75.89          & 73.96          & 42.02          & 38.98          & 45.07          & 76.27          & 77.48          & 46.24          & 77.50          & 77.11          & 78.67          & 58.35                          \\
                                       & RFNet                      & 48.87          & 80.51          & 39.40          & 41.63          & 81.76          & 81.62          & 49.22          & 51.44          & 55.80          & 82.70          & 84.20          & 57.05          & 82.47          & 83.08          & 83.59          & 66.89                          \\
                                       & mmFormer                   & 46.59          & 79.84          & 39.05          & 41.98          & 82.35          & 82.79          & 47.09          & 49.67          & 52.50          & 82.90          & 83.35          & 54.39          & 82.88          & 83.23          & 83.08          & 66.11                          \\
                                       & M$^{3}$AE                  & \second{58.68} & \third{83.95}  & \second{54.00} & \second{57.09} & \third{84.79}  & 85.12          & \second{60.90} & \second{60.25} & \second{63.03} & \second{84.90} & \third{85.80}  & \second{63.50} & \second{85.43} & 85.23          & \third{85.74}  & \second{73.23}                 \\
                                       & M$^{2}$FTrans              & 54.58          & \second{84.04} & 45.79          & 49.81          & \second{85.03} & \second{85.51} & 55.34          & 56.60          & 60.18          & \third{84.87}  & 85.56          & 60.38          & \third{84.90}  & \second{85.64} & \second{85.78} & 70.93                          \\
                                       & LS3M                       & 52.15          & 83.36          & 46.25          & 52.38          & 84.78          & \third{85.37}  & 56.45          & 56.76          & 60.58          & 84.57          & \second{86.44} & 61.12          & 84.71          & 85.35          & 85.53          & 71.05                          \\
                                       & IM-Fuse                    & 51.74          & 82.85          & 48.80          & 51.74          & 84.40          & 84.30          & 56.04          & 56.68          & 58.71          & 84.51          & 85.34          & 60.94          & 84.25          & 84.81          & 84.90          & 70.67                          \\
                                       & M$^{2}$SegMamba            & \third{54.84}  & 81.72          & \third{51.09}  & \third{55.24}  & 84.55          & 83.49          & \third{59.48}  & \third{58.69}  & \third{60.59}  & 84.53          & 84.67          & \third{62.37}  & 84.27          & \third{85.40}  & 85.03          & \third{71.73}                  \\
                                       & Ours                       & \first{61.12}  & \first{86.95}  & \first{55.35}  & \first{60.00}  & \first{87.54}  & \first{87.40}  & \first{63.15}  & \first{62.11}  & \first{65.11}  & \first{87.59}  & \first{87.94}  & \first{65.93}  & \first{87.80}  & \first{87.87}  & \first{87.98}  & \first{75.59}                  \\
        \bottomrule
    \end{tabular}
    \caption{\textbf{Performance comparison (DSC~\%) on BraTS 2023.} Available and missing modalities are denoted by \filledcirc~and \opencirc, respectively. \legendsquare{colorbest}, \legendsquare{colorsecond}, and \legendsquare{colorthird} indicate the first-, second-, and third-best performance, respectively. WT, TC, and ET represent whole tumor, tumor core, and enhancing tumor, respectively.}
    \label{tab:dsc23}
\end{table*}

\begin{table*}[htbp]
    \centering
    \scriptsize
    \renewcommand{\arraystretch}{1}
    \setlength{\tabcolsep}{4.3pt}
    \begin{tabular}{c|c|*{15}{c}|c}
        \toprule
        \multirow{4}{*}{\textbf{Type}} & \multicolumn{1}{c|}{Flair} & \opencirc      & \opencirc      & \opencirc      & \filledcirc    & \opencirc      & \opencirc      & \filledcirc    & \opencirc      & \filledcirc    & \filledcirc    & \filledcirc    & \filledcirc    & \filledcirc    & \opencirc      & \filledcirc    & \multirow{4}{*}{\textbf{Avg.}} \\
                                       & T1                         & \opencirc      & \opencirc      & \filledcirc    & \opencirc      & \opencirc      & \filledcirc    & \filledcirc    & \filledcirc    & \opencirc      & \opencirc      & \filledcirc    & \filledcirc    & \opencirc      & \filledcirc    & \filledcirc    &                                \\
                                       & T1ce                       & \opencirc      & \filledcirc    & \opencirc      & \opencirc      & \filledcirc    & \filledcirc    & \opencirc      & \opencirc      & \opencirc      & \filledcirc    & \filledcirc    & \opencirc      & \filledcirc    & \filledcirc    & \filledcirc    &                                \\
                                       & T2                         & \filledcirc    & \opencirc      & \opencirc      & \opencirc      & \filledcirc    & \opencirc      & \opencirc      & \filledcirc    & \filledcirc    & \opencirc      & \opencirc      & \filledcirc    & \filledcirc    & \filledcirc    & \filledcirc    &                                \\
        \midrule
        \multirow{10}{*}{\textbf{WT}}  & HeMIS                      & 52.37          & 33.85          & 36.56          & 60.68          & 56.90          & 44.92          & 65.21          & 58.29          & 67.52          & 65.62          & 67.85          & 69.08          & 69.73          & 60.35          & 70.10          & 58.60                          \\
                                       & U-HEVD                     & 71.74          & 56.95          & 68.75          & 77.68          & 76.24          & 70.47          & 83.39          & 78.48          & 83.00          & 81.95          & 84.10          & 84.62          & 84.07          & 79.07          & 85.01          & 77.70                          \\
                                       & RFNet                      & 81.65          & 76.66          & 77.19          & 85.71          & 83.77          & 79.45          & 87.36          & 83.35          & 87.65          & 88.10          & 88.21          & 88.02          & 88.69          & 84.22          & 88.66          & 84.58                          \\
                                       & mmFormer                   & 81.81          & 77.10          & 77.80          & 85.66          & 84.00          & 79.85          & 87.52          & 83.62          & 87.72          & 88.12          & 88.37          & 88.12          & 88.64          & 84.48          & 88.65          & 84.76                          \\
                                       & M$^{3}$AE                  & \second{83.28} & \second{79.63} & \second{80.19} & \second{87.44} & \third{85.25}  & 81.24          & \second{88.91} & \third{84.49}  & \third{89.07}  & \second{89.43} & \second{89.52} & \third{89.30}  & 89.66          & 85.26          & \second{90.02} & \second{86.18}                 \\
                                       & M$^{2}$FTrans              & 81.94          & 78.36          & 78.85          & \third{86.90}  & 84.33          & 80.60          & 88.09          & 83.92          & 88.23          & 88.79          & 88.99          & 88.51          & 89.11          & 84.75          & 89.14          & 85.37                          \\
                                       & LS3M                       & 82.44          & 78.48          & \third{79.08}  & 86.19          & 84.70          & 81.11          & 87.85          & 84.31          & 88.34          & 88.37          & 88.65          & 88.73          & 89.12          & 85.13          & 89.19          & 85.45                          \\
                                       & IM-Fuse                    & \third{82.61}  & \third{79.42}  & 78.96          & 86.36          & \second{85.33} & \second{81.89} & \third{88.45}  & \second{84.62} & \second{89.08} & 89.11          & 89.40          & \second{89.50} & \second{89.91} & \second{85.64} & \third{89.98}  & \third{86.02}                  \\
                                       & M$^{2}$SegMamba            & 82.24          & 79.10          & 78.66          & 85.90          & 84.97          & \third{81.35}  & 88.23          & 84.02          & 88.55          & \third{89.17}  & \third{89.43}  & 88.80          & \third{89.74}  & \third{85.37}  & 89.77          & 85.69                          \\
                                       & Ours                       & \first{85.37}  & \first{82.73}  & \first{82.75}  & \first{89.00}  & \first{87.06}  & \first{84.19}  & \first{90.34}  & \first{86.63}  & \first{90.44}  & \first{90.79}  & \first{90.84}  & \first{90.65}  & \first{91.18}  & \first{87.22}  & \first{91.14}  & \first{88.02}                  \\
        \hline
        \multirow{10}{*}{\textbf{TC}}  & HeMIS                      & 18.78          & 27.55          & 10.98          & 3.55           & 42.84          & 36.15          & 20.56          & 27.16          & 31.27          & 42.18          & 42.91          & 29.16          & 48.56          & 42.43          & 46.47          & 31.37                          \\
                                       & U-HEVD                     & 45.15          & 56.49          & 50.17          & 41.18          & 66.05          & 64.83          & 56.11          & 56.43          & 56.01          & 69.23          & 71.78          & 59.61          & 71.91          & 69.86          & 73.27          & 60.54                          \\
                                       & RFNet                      & 61.90          & 74.31          & 60.43          & 57.53          & 77.39          & 76.88          & 65.32          & 65.29          & 65.73          & 78.46          & 79.35          & 67.21          & 79.67          & 78.43          & 80.07          & 71.20                          \\
                                       & mmFormer                   & 63.08          & 74.14          & 60.55          & 57.46          & 78.12          & 77.08          & 66.50          & 67.66          & 68.16          & 78.42          & 80.25          & 69.60          & 80.47          & 79.86          & 81.59          & 72.20                          \\
                                       & M$^{3}$AE                  & \second{68.33} & 76.12          & \second{66.40} & \second{63.14} & \second{81.66} & \second{80.06} & \second{69.27} & \second{70.18} & \second{71.32} & \third{80.65}  & \second{83.21} & \second{71.87} & \second{83.64} & \second{82.38} & \second{84.68} & \second{75.53}                 \\
                                       & M$^{2}$FTrans              & 62.26          & 75.89          & 63.15          & 60.66          & 79.22          & 79.10          & 67.14          & 66.75          & 67.55          & 80.20          & 81.43          & 69.32          & 81.33          & 80.23          & 81.89          & 73.07                          \\
                                       & LS3M                       & 63.51          & 75.84          & 62.18          & 59.88          & 78.95          & 78.68          & 67.67          & 68.28          & 69.17          & 79.22          & 80.83          & 71.04          & 80.79          & 80.26          & 81.72          & 73.20                          \\
                                       & IM-Fuse                    & \third{65.28}  & \third{76.17}  & 63.36          & 58.42          & 79.75          & 79.27          & 68.18          & 68.76          & 69.67          & 80.62          & 82.18          & \third{71.80}  & \third{82.88}  & 80.39          & 82.86          & 73.97                          \\
                                       & M$^{2}$SegMamba            & 64.57          & \second{77.33} & \third{64.65}  & \third{61.43}  & \third{80.77}  & \third{79.83}  & \third{69.01}  & \third{68.95}  & \third{70.44}  & \second{81.47} & \third{82.92}  & 71.58          & 82.09          & \third{81.60}  & \third{83.73}  & \third{74.69}                  \\
                                       & Ours                       & \first{68.65}  & \first{79.68}  & \first{68.75}  & \first{66.13}  & \first{82.74}  & \first{81.69}  & \first{72.83}  & \first{73.14}  & \first{73.51}  & \first{83.14}  & \first{83.84}  & \first{75.09}  & \first{85.05}  & \first{83.03}  & \first{85.08}  & \first{77.49}                  \\
        \hline
        \multirow{10}{*}{\textbf{ET}}  & HeMIS                      & 16.08          & 26.47          & 8.62           & 2.15           & 41.52          & 34.96          & 15.59          & 24.06          & 26.38          & 40.02          & 41.51          & 24.71          & 46.57          & 41.10          & 44.98          & 28.98                          \\
                                       & U-HEVD                     & 39.93          & 54.21          & 47.12          & 34.97          & 63.24          & 62.44          & 49.81          & 51.74          & 46.06          & 67.22          & 69.93          & 52.31          & 69.06          & 67.12          & 70.79          & 56.40                          \\
                                       & RFNet                      & 57.23          & 73.53          & 57.04          & 53.70          & 75.06          & 76.25          & 61.37          & 60.60          & 61.18          & 77.14          & 78.59          & 63.73          & 77.67          & 76.20          & 78.75          & 68.54                          \\
                                       & mmFormer                   & 58.54          & 73.29          & 56.63          & 53.53          & 77.49          & 76.55          & 61.92          & 63.25          & 62.57          & 77.17          & 78.92          & 65.42          & 79.18          & 78.64          & 80.24          & 69.56                          \\
                                       & M$^{3}$AE                  & \second{63.55} & 73.98          & \second{61.92} & \second{60.08} & \second{79.04} & 77.53          & \third{64.91}  & \second{65.46} & \second{66.89} & 78.30          & 80.81          & \third{67.34}  & \second{81.03} & \second{79.82} & \second{82.12} & \second{72.19}                 \\
                                       & M$^{2}$FTrans              & 58.47          & 75.19          & 58.51          & 57.33          & 78.23          & 77.71          & 63.51          & 63.16          & 63.76          & \second{79.48} & \third{80.83}  & 65.97          & \third{80.36}  & 79.31          & 81.25          & 70.87                          \\
                                       & LS3M                       & 59.17          & \third{75.79}  & 59.54          & 56.47          & 77.51          & \second{78.44} & 64.71          & 63.79          & 63.77          & 79.05          & \second{80.83} & 66.42          & 79.29          & 78.19          & 80.21          & 70.88                          \\
                                       & IM-Fuse                    & \third{61.36}  & \second{76.02} & \third{61.01}  & 55.87          & 77.99          & \third{78.33}  & 64.59          & \third{64.87}  & 65.06          & \third{79.28}  & 80.43          & 67.08          & 80.01          & 78.94          & 80.77          & 71.44                          \\
                                       & M$^{2}$SegMamba            & 60.66          & 75.19          & 60.81          & \third{58.50}  & \third{78.51}  & 77.79          & \second{65.04} & 64.53          & \third{65.59}  & 79.24          & 80.56          & \second{67.66} & 80.32          & \third{79.44}  & \third{81.33}  & \third{71.68}                  \\
                                       & Ours                       & \first{66.00}  & \first{78.57}  & \first{66.04}  & \first{64.40}  & \first{80.99}  & \first{80.34}  & \first{69.58}  & \first{69.43}  & \first{69.73}  & \first{81.75}  & \first{82.41}  & \first{70.89}  & \first{82.30}  & \first{81.58}  & \first{82.81}  & \first{75.12}                  \\
        \bottomrule
    \end{tabular}
    \caption{\textbf{Performance comparison (DSC~\%) on BraTS 2024.} Available and missing modalities are denoted by \filledcirc~and \opencirc, respectively. \legendsquare{colorbest}, \legendsquare{colorsecond}, and \legendsquare{colorthird} indicate the first-, second-, and third-best performance, respectively. WT, TC, and ET represent whole tumor, tumor core, and enhancing tumor, respectively.}
    \label{tab:dsc24}
\end{table*}

\subsection{Ablation Study}\label{sec:ablation_study}
We conduct ablations on the two-stage heterogeneous design and key hyperparameters of the proposed UniME using the BraTS 2023 dataset.

\begin{figure}[H]
    \centering
    \includegraphics[width=0.85\linewidth]{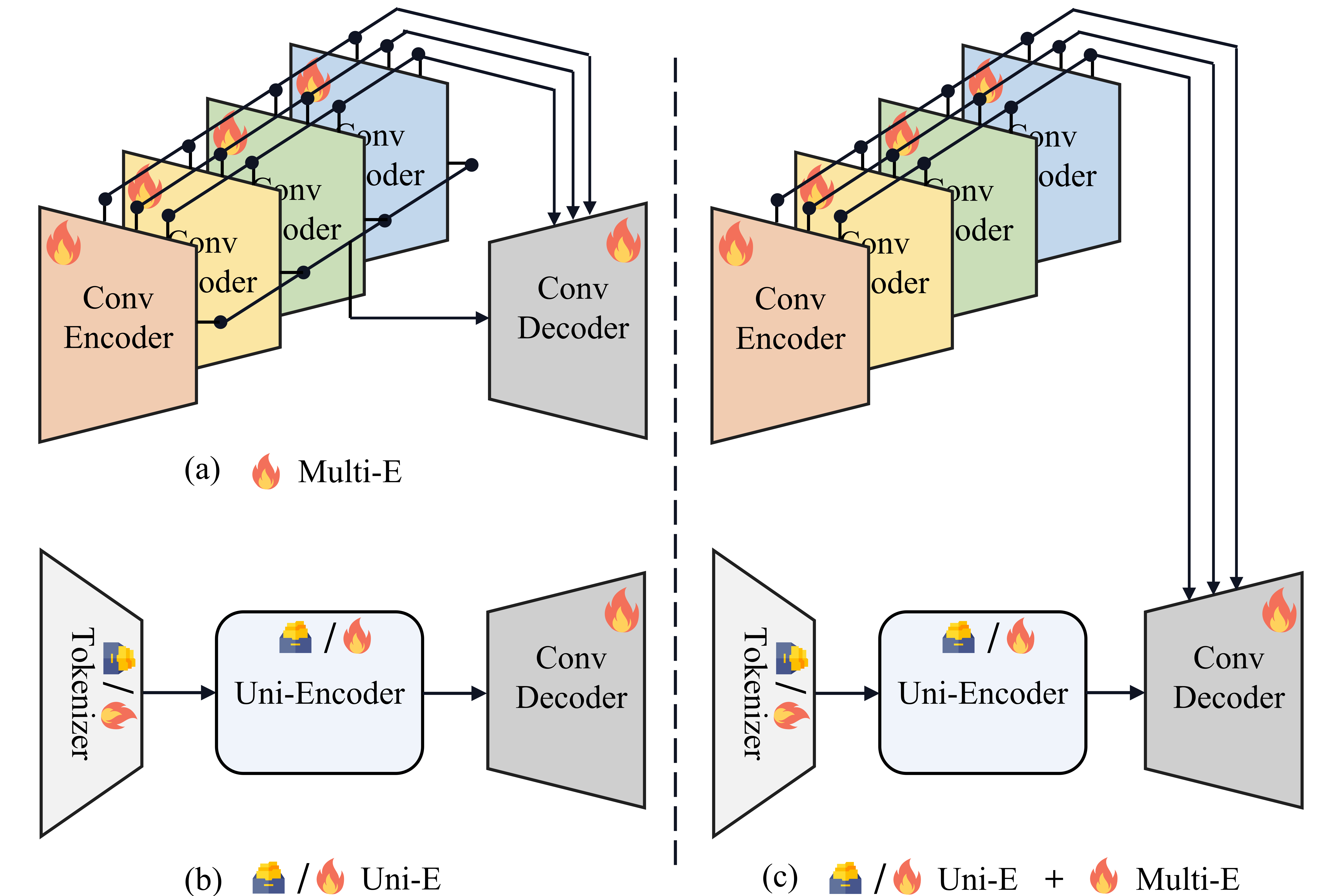}
    \caption{\textbf{Architectural designs corresponding to ablation study in Tab.~\ref{tab:ablation_study1}}. \includegraphics[height=1em]{archive.png} / \includegraphics[height=1em]{flame.png} denote modules initialized from pretrained weights and trained from scratch, respectively.}
    \label{fig:ablation_architecture}
\end{figure}

\noindent \textbf{Effect of Two-Stage Heterogeneous Design.}~~Fig.~\ref{fig:ablation_architecture} shows the architectural variations tested in Tab.~\ref{tab:ablation_study1}. Without pretraining, the Multi-Encoders architecture combined with a standard CNN decoder achieves a mean DSC of 78.90. A randomly initialized Uni-Encoder alone performs worse. However, combining Uni-Encoder and Multi-Encoders architectures improves performance to a mean DSC of 81.22, demonstrating their complementary strengths in modeling cross-modal and modality-specific features. Masked self-supervised pretraining with a single Uni-Encoder achieves a mean DSC of 82.45, notably enhancing ET segmentation. Integrating both pretraining and heterogeneous design results in the best performance, reaching a mean DSC of 83.49.

\noindent\textbf{Effect of Uni-Encoder Scale.}~~Tab.~\ref{tab:ablation_study2} presents the performance across Uni-Encoder scales. A small scale Uni-Encoder achieves the lowest mean DSC of 80.00, which indicates limited ability for modeling complex multimodal features. Conversely, a larger encoder does not yield consistent gains. A moderate scale provides a balance between accuracy and efficiency, achieving the best mean DSC of 83.49.

\noindent \textbf{Effect of Hyperparameters in Stage 1.}~~{
\parfillskip=0pt
In Tab.~\ref{tab:ablation_study3} and Tab.~\ref{tab:ablation_study4}, we analyze the patch masking ratio and the number of register tokens during Stage 1. Performance reaches its maximum at $q_m$ = 75\% and 4 register tokens, respectively. Both parameters initially improve but subsequently degrade after reaching an optimum. This is because a moderate masking ratio
\par
}

\begin{table}[H]
    \centering
    \scriptsize
    \scalebox{0.9}{
        \begin{tabular}{ccc|ccc|c}
            \toprule
            \multicolumn{3}{c|}{\textbf{Components}}       & \multicolumn{3}{c|}{\textbf{Average Region DSC (\%)}} & \multicolumn{1}{c}{\textbf{Overall}}                                                                                       \\
            \cmidrule(lr){1-3} \cmidrule(lr){4-6} \cmidrule(l){7-7}
            \includegraphics[height=1em]{archive.png}Uni-E & \includegraphics[height=1em]{flame.png}Uni-E          & \includegraphics[height=1em]{flame.png}Multi-E & WT               & TC               & ET               & Mean             \\
            \midrule
            \opencirc                                      & \opencirc                                             & \filledcirc                                    & 88.25            & 79.62            & 68.83            & 78.90            \\
            \opencirc                                      & \filledcirc                                           & \opencirc                                      & 87.86            & 77.43            & 66.46            & 77.25            \\
            \filledcirc                                    & \opencirc                                             & \opencirc                                      & \absecond{90.29} & \absecond{83.39} & \absecond{73.67} & \absecond{82.45} \\
            \opencirc                                      & \filledcirc                                           & \filledcirc                                    & \abthird{88.95}  & \abthird{82.09}  & \abthird{72.61}  & \abthird{81.22}  \\
            \filledcirc                                    & \opencirc                                             & \filledcirc                                    & \abfirst{90.38}  & \abfirst{84.51}  & \abfirst{75.59}  & \abfirst{83.49}  \\
            \bottomrule
        \end{tabular}
    }
    \caption{\textbf{Ablation on two-stage heterogeneous design.} \legendsquare{ablationcolorbest}, \legendsquare{ablationcolorsecond}, and \legendsquare{ablationcolorthird} indicate the first, second, and third best results, respectively. Uni-E and Multi-E denote Uni-Encoder and Multi-Encoders, respectively. \includegraphics[height=1em]{archive.png} / \includegraphics[height=1em]{flame.png} denote modules initialized from pretrained weights and trained from scratch, respectively.
    }
    \label{tab:ablation_study1}
\end{table}

\begin{table}[H]
    \centering
    \scriptsize
    \scalebox{0.9}{
        \begin{tabular}{c|ccc|ccc|c}
            \toprule
            \multirow{2.3}{*}{\textbf{Scale}} & \multicolumn{3}{c|}{\textbf{Configuration}} & \multicolumn{3}{c|}{\textbf{Average Region DSC (\%)}} & \textbf{Overall}                                                                               \\
            \cmidrule(lr){2-4} \cmidrule(lr){5-7} \cmidrule(l){8-8}
                                              & Heads                                       & $L$                                                   & $d_{\text{embed}}$ & WT               & TC               & ET               & Mean             \\
            \midrule
            Small                             & 12                                          & 12                                                    & 864                & \abthird{89.50}  & \abthird{80.77}  & \abthird{69.72}  & \abthird{80.00}  \\
            Base                              & 12                                          & 16                                                    & 864                & \absecond{90.38} & \abfirst{84.51}  & \abfirst{75.59}  & \abfirst{83.49}  \\
            Large                             & 16                                          & 24                                                    & 1056               & \abfirst{90.50}  & \absecond{84.09} & \absecond{75.08} & \absecond{83.22} \\
            \bottomrule
        \end{tabular}
    }
    \caption{\textbf{Ablation on scale of Uni-Encoder.} \legendsquare{ablationcolorbest}, \legendsquare{ablationcolorsecond}, and \legendsquare{ablationcolorthird} indicate the first, second, and third best results, respectively.}
    \label{tab:ablation_study2}
\end{table}

\begin{table}[H]
    \centering
    \scriptsize
    \scalebox{0.9}{
        \begin{tabular}{c|ccc|c}
            \toprule
            \multirow{2.3}{*}{\textbf{$q_{m}$}} & \multicolumn{3}{c|}{\textbf{Average Region DSC (\%)}} & \multicolumn{1}{c}{\textbf{Overall}}                                       \\
            \cmidrule(lr){2-4} \cmidrule(l){5-5}
                                                & WT                                                    & TC                                   & ET               & Mean             \\
            \midrule
            25.0~\%                             & 90.24                                                 & 83.51                                & 74.71            & 82.82            \\
            50.0~\%                             & 90.20                                                 & \absecond{84.15}                     & \abthird{75.05}  & \abthird{83.13}  \\
            67.5~\%                             & \abfirst{90.44}                                       & \abthird{84.10}                      & \absecond{75.13} & \absecond{83.22} \\
            75.0~\%                             & \abthird{90.38}                                       & \abfirst{84.51}                      & \abfirst{75.59}  & \abfirst{83.49}  \\
            87.5~\%                             & \absecond{90.41}                                      & 83.51                                & 74.96            & 82.96            \\
            \bottomrule
        \end{tabular}
    }
    \caption{\textbf{Ablation on patch masking ratio $q_{m}$}. \legendsquare{ablationcolorbest}, \legendsquare{ablationcolorsecond}, and \legendsquare{ablationcolorthird} indicate the first, second, and third best results, respectively.}
    \label{tab:ablation_study3}
\end{table}

\noindent strengthens cross-modal semantic learning, while an excessively high ratio removes too much information, which impairs training. Moreover, MRI volumes contain more complex background structures, requiring careful consideration: too few register tokens miss salient context, whereas too many introduce redundancy and reduce training efficiency.

\begin{table}[t]
    \centering
    \scriptsize
    \scalebox{0.9}{
        \begin{tabular}{c|ccc|c}
            \toprule
            \multirow{2.3}{*}{\textbf{\# Register Tokens}} & \multicolumn{3}{c|}{\textbf{Average Region DSC (\%)}} & \multicolumn{1}{c}{\textbf{Overall}}                                       \\
            \cmidrule(lr){2-4} \cmidrule(l){5-5}
                                                           & WT                                                    & TC                                   & ET               & Mean             \\
            \midrule
            0                                              & 89.88                                                 & 83.64                                & 74.39            & 82.64            \\
            1                                              & 90.34                                                 & \abthird{83.82}                      & \absecond{75.02} & \absecond{83.06} \\
            2                                              & \absecond{90.40}                                      & \absecond{83.86}                     & \abthird{74.59}  & \abthird{82.95}  \\
            4                                              & \abthird{90.38}                                       & \abfirst{84.51}                      & \abfirst{75.59}  & \abfirst{83.49}  \\
            8                                              & \abfirst{90.52}                                       & 83.71                                & 74.55            & 82.93            \\
            \bottomrule
        \end{tabular}
    }
    \caption{\textbf{Ablation on the number of register tokens}. \legendsquare{ablationcolorbest}, \legendsquare{ablationcolorsecond}, and \legendsquare{ablationcolorthird} indicate the first, second, and third best results, respectively.}
    \label{tab:ablation_study4}
\end{table}

\noindent \textbf{Effect of LLRD Strategy in Stage 2.}~~Tab.~\ref{tab:ablation_study5} shows the impact of different LLRD rates $\omega$ and of weight freezing applied to the Uni-Encoder during Stage 2. The LLRD strategy outperforms weight freezing, with optimal performance at the decay rate $\omega = 0.75$. Freezing weights or using too small a decay limits adaptation to the target task, whereas an overly large decay fails to preserve pretrained semantics and undermines the purpose of the LLRD strategy.
\begin{table}[htbp]
    \centering
    \scriptsize
    \scalebox{0.9}{
        \begin{tabular}{c|ccc|c}
            \toprule
            \multirow{2.3}{*}{\textbf{$\omega$}}        & \multicolumn{3}{c|}{\textbf{Average Region DSC (\%)}} & \multicolumn{1}{c}{\textbf{Overall}}                                       \\
            \cmidrule(lr){2-4} \cmidrule(l){5-5}
                                                        & WT                                                    & TC                                   & ET               & Mean             \\
            \midrule
            \includegraphics[height=1em]{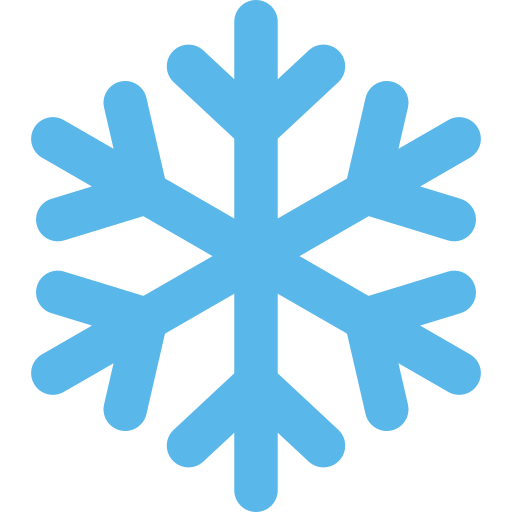} & 89.56                                                 & 82.91                                & 74.12            & 82.20            \\
            0.65                                        & 90.29                                                 & \abthird{84.20}                      & \absecond{75.53} & \absecond{83.34} \\
            0.75                                        & \absecond{90.38}                                      & \abfirst{84.51}                      & \abfirst{75.59}  & \abfirst{83.49}  \\
            0.85                                        & \abthird{90.29}                                       & \absecond{84.24}                     & 74.88            & \abthird{83.14}  \\
            1.00                                        & \abfirst{90.39}                                       & 83.48                                & \abthird{74.96}  & 82.94            \\
            \bottomrule
        \end{tabular}
    }
    \caption{\textbf{Ablation on LLRD rates}. \legendsquare{ablationcolorbest}, \legendsquare{ablationcolorsecond}, and \legendsquare{ablationcolorthird} indicate the first, second, and third best results, respectively. \includegraphics[height=0.9em]{snowflake.png} denotes that the Uni-Encoder weights are frozen during fine-tuning.}
    \label{tab:ablation_study5}
\end{table}
\section{Conclusion}
In this work, we introduce UniME, a two-stage heterogeneous method for incomplete modality brain tumor segmentation. It reconciles the trade-off among fine-grained structure capture, cross-modal complementarity modeling, and effective exploitation of available modalities. UniME first pretrains a Uni-Encoder with masked self-supervision to obtain unified representations robust to absent modalities. Then, it introduces Multi-Encoders to enrich the Uni-Encoder with multi-scale, fine-grained features. Experiments on BraTS 2023 and BraTS 2024 datasets demonstrate that UniME significantly outperforms existing state-of-the-art methods.

Our findings highlight the two-stage heterogeneous framework’s potential for medical image segmentation and its adaptability to challenging cases with missing modalities.

\section*{Acknowledgement}
This study was supported in part by the National Natural Science Foundation of China (12426310, 62372269, 12371492), in part by the National Key Research and Development Program (2021YFA1000202), in part by the Key Technology Research and Development Program of Shandong (2024TSGC1132), in part by the Shandong Province Natural Science Foundation (ZR2025MS70, ZR2022MF245).

    {
        \small
        \bibliographystyle{ieeenat_fullname}
        \bibliography{main}
    }

\clearpage

\end{document}